\documentclass[letterpaper, 10 pt, conference]{IEEEconf}  

\IEEEoverridecommandlockouts                              

\usepackage{amsmath,amssymb,amsfonts}
\usepackage{algorithm}
\usepackage{algpseudocode}
\usepackage{subfig}
\usepackage{graphicx}
\usepackage{textcomp}
\usepackage{xcolor}
\usepackage{todonotes}
\usepackage{soul}
\usepackage[utf8]{inputenc}
\usepackage{booktabs}
\usepackage{hyperref}
\usepackage[flushleft]{threeparttable}

\def\BibTeX{{\rm B\kern-.05em{\sc i\kern-.025em b}\kern-.08em
    T\kern-.1667em\lower.7ex\hbox{E}\kern-.125emX}}

\begin{document}

\title{\LARGE \bf
I Move Therefore I Learn:\\Experience-Based Traversability in Outdoor Robotics
}

\author{Miguel \'Angel de Miguel$^{1}$, Jorge Beltr\'an$^{1}$, Juan S. Cely$^{1}$, Francisco Mart\'in$^{1}$,\\
Juan Carlos Manzanares$^{1}$, and Alberto Garc\'ia$^{1}$
\thanks{*This work is partially funded under Project PID2021-126592OB-C22 funded by MCIN/AEI/10.13039/501100011033, the grant TED2021-132356B-I00 funded by MCIN/AEI/10.13039/501100011033, the GETROPEX project F1269 - 2025/00014/033 with funding by self program URJC and by the “European Union NextGenerationEU/PRTR", and by CORESENSE project with funding from the European Union’s Horizon Europe Research and Innovation Programme (Grant Agreement No. 101070254)}%
\thanks{$^{1}$Intelligent Robotics Lab, Universidad Rey Juan Carlos, 28943, Fuenlabrada, Spain 
{\footnotesize\ttfamily \{miguelangel.demiguel, jorge.beltran,  juan.cely, francisco.rico, juancarlos.serrano, alberto.gomezjacinto\}@urjc.es}}
\thanks{%
    \textbf{Author contributions —}
    \textbf{Conceptualization}: F.M.; %
    \textbf{Data curation}: J.B., J.C.M., A.G.; %
    \textbf{Formal analysis}: M.A.d.M., J.B., J.S.C.; %
    \textbf{Investigation}: M.A.d.M., J.B., J.S.C, F.M.; %
    \textbf{Methodology}: M.A.d.M., J.B., J.S.C; %
    \textbf{Software}: M.A.d.M., F.M.; %
    \textbf{Validation}: M.A.d.M., J.B., J.S.C.; %
    \textbf{Writing–original draft}: M.A.d.M., J.B, J.S.C.; %
  }
}

\maketitle

\begin{abstract}

Accurate traversability estimation is essential for safe and effective navigation of outdoor robots operating in complex environments. This paper introduces a novel experience-based method that allows robots to autonomously learn which terrains are traversable based on prior navigation experience, without relying on extensive pre-labeled datasets. The approach integrates elevation and texture data into multi-layered grid maps, which are processed using a variational autoencoder (VAE) trained on a generic texture dataset. During an initial teleoperated phase, the robot collects sensory data while moving around the environment. These experiences are encoded into compact feature vectors and clustered using the BIRCH algorithm to represent traversable terrain areas efficiently. In deployment, the robot compares new terrain patches to its learned feature clusters to assess traversability in real time. The proposed method does not require training with data from the targeted scenarios, generalizes across diverse surfaces and platforms, and dynamically adapts as new terrains are encountered. Extensive evaluations on both synthetic benchmarks and real-world scenarios with wheeled and legged robots demonstrate its effectiveness, robustness, and superior adaptability compared to state-of-the-art approaches.
\end{abstract}



\section{Introduction}
Field robotics is a research area focused on deploying robots in outdoor environments, particularly in spaces with minimal or no structural organization. In contrast to highly structured indoor settings, outdoor environments present challenges such as uneven terrain, the absence of consistent, perpendicular surfaces, and human-made regulations that impose access restrictions.
For instance, streets may require sidewalk navigation, with crossing only allowed at designated points, and slopes might necessitate detours. Moreover, certain surfaces may appear as obstacles merely due to a robot’s inclination and elevation perception. Navigating outdoor environments is inherently more complex than indoor navigation, with one advantage being the availability of global positioning through GPS. However, even then, geolocation must be related with pre-existing maps and enhanced by other sensor data for precise navigation.

\begin{figure}[!thb]
  \centering
  \includegraphics[width=\linewidth]{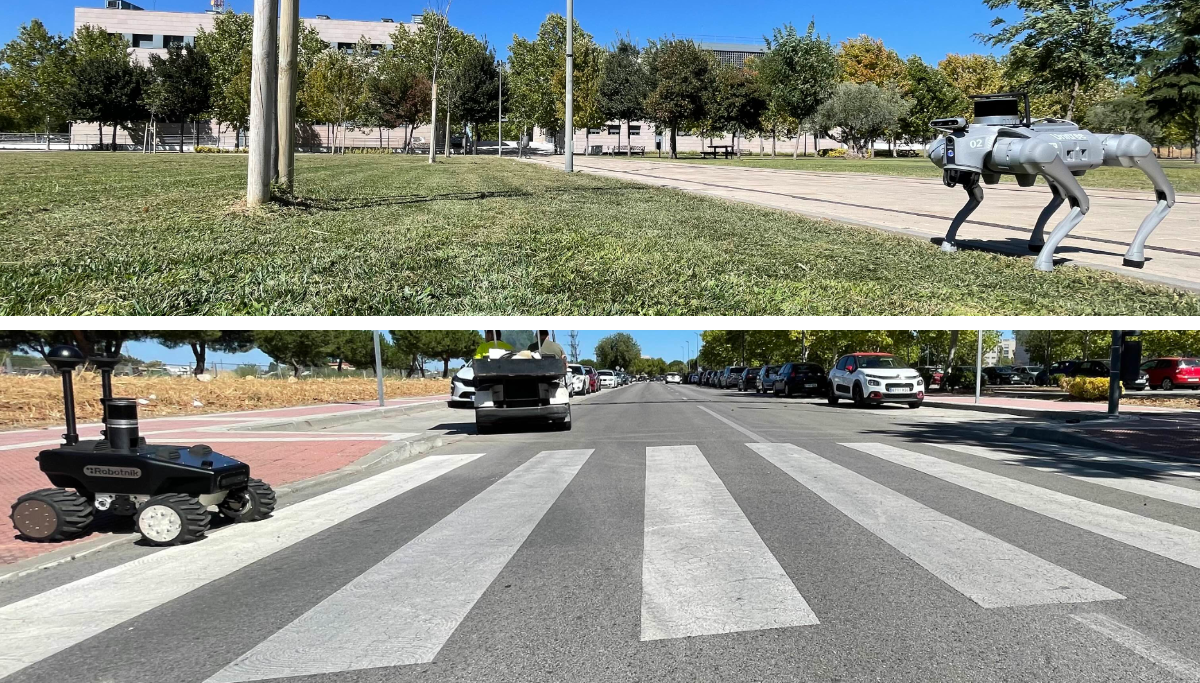}
  \caption{Robots and scenarios used for the experimental validation of this work: Quadruped robot Go2 in a paviment-grass scenario (upper image) and wheeled robot Summit XL in a pedestrian crossing (bottom image).}
  \label{fig:robots}
\end{figure}

A key capability for outdoor robots is traversability, or the ability to move safely through an area. This depends on factors like surface inclination, material composition, obstacles, and regulations. Steeper or debris-covered terrain becomes more difficult, and traversability is also influenced by the robot’s locomotion system, power, and design.

In this work, we contribute to the advancement of outdoor navigation by proposing a method to identify areas that a robot can safely traverse. Rather than relying on a roboticist to meticulously define the conditions under which a terrain is considered traversable based on pre-established criteria, our approach allows the robot to learn which areas are traversable through experience. This enables the robot to quickly adapt to new environments, facilitating smoother and more efficient deployment in unfamiliar settings. 

The core concept of our work follows an emerging trend in the field: \emph{a robot can traverse a location because it has successfully done so before on a terrain with similar characteristics}. Our approximation relies on the existence of an exploration phase where a teleoperator guides the robot across a given environment. This stage allows the extraction of distinctive terrain descriptors of the navigated areas, allowing the robot to generalize this knowledge to determine the traversability of visited regions in the future. The learning process occurs in real-time; if the robot initially deems an area untraversable but is later guided through it by an operator, it will adjust its understanding and expand the set of traversable areas to include surfaces with similar characteristics.

The presented approach was validated in both synthetic and real-world settings. On the one hand, the well-known NEGS-UGV benchmark \cite{sanchez_automatically_2022} was used to test its performance against perfect ground truth annotations. On the other hand, experiments with two heterogeneous robotic platforms on our university campus were conducted to further assess the capabilities of the method under challenging scenarios with real sensor data, as shown in Fig. \ref{fig:robots}.

In summary, our contributions to the field are as follows:
\begin{itemize}
    \item A novel \textbf{environment encoding technique} that enables the determination of traversability.
    \item A \textbf{mathematical framework for} summarizing map patch information and comparing patches to calculate distances, which is key to \textbf{determining the traversability of the environment}.
    \item  A \textbf{deployment methodology} for learning traversable zones through online robot exploration.
    \item An \textbf{open-source ROS 2 implementation}, compatible with established navigation frameworks such as Nav2.
\end{itemize}



\section{Related Works}
\label{sec:related-works}

In this section, we review the most relevant works that have informed our research.
In particular, we build upon the traversability concept outlined by Benrabah et al. \cite{benrabah_review_2024}, distinguishing terrains that merely allow passage from those enabling efficient locomotion, and review relevant methods that leverage images, LiDAR, terrain interaction, simulation, and prior experiential learning. We will also delve into the inherently complex challenge of selecting and evaluating the most commonly used datasets to validate these methods.

The most intuitive method for assessing terrain suitability relies on visual information, using images to determine whether a surface is traversable and to generate paths or waypoints accordingly as denoted by \cite{frey_fast_2023}. 
Nonetheless, vision-based terrain classification methods are significantly dependent on accurate texture recognition, a challenge extensively explored by \cite{cimpoi_describing_2014}. 
Particularly, in urban environments, the diversity of terrain types complicates this task, as demonstrated by \cite{hyun_street_2022}. In unfamiliar environments, where terrain classes cannot be reliably identified, \cite{seo_learning_2023} propose integrating prior experience and labeled data to enhance the robustness of the method.

Complementing visual approaches, LiDAR-based methods offer robust alternatives for traversability estimation. The work in \cite{agishev_monoforce_2024} propose a method that detects terrain deformation through interaction with the ground, classifying it as traversable or not, similar situation is described by \cite{lim_similar_2024}.

Beyond perception, in unstructured and unknown terrains, it is common for systems to output semantic labels that guide high-level control strategies in subsequent iterations. For instance, one approach generates Model Predictive Control (MPC) strategies based on semantic information derived from sensor data \cite{han_model_2024}. In a related context, autonomous excavators have been developed for construction environments with unstructured terrain, leveraging both images and LiDAR point clouds as inputs to a terrain classification decoder \cite{guan_tns_2022}.

When locomotion is performed by a legged robot, footstep interactions can provide valuable feedback to the system regarding terrain characteristics, enabling the classification of surfaces as traversable or non-traversable. This approach has been explored by several authors, including 
\cite{yu_semi-supervised_2022}, 
\cite{zhang_learning_2024}, 
\cite{miki_learning_2022},
and \cite{jenelten_perceptive_2020}.
Similarly, and building on the integration of simulated and real data, \cite{wellhausen_where_2019} introduce a system that correlates visual input with footstep data to generate optimal paths over traversable terrain. 

Extending this integration of simulated and real-world data, and given the wide variety of terrain types, simulation-based learning offers a promising solution. Authors in \cite{chen_identifying_2024} propose a method that combines real-world imagery with simulated data to train a decoder capable of identifying terrain classes. Their approach was validated using a quadrupedal robot. Similarly, \cite{chavez-garcia_learning_2018} focus on learning heightmaps from simulated images and validating the results in real-world scenarios.

In parallel with these approaches, machine learning algorithms have significantly changed the requirements for map and terrain data. In \cite{castro_how_2023}, it is demonstrated that learning from costmap representations is feasible when proprioceptive and exteroceptive data are combined, validating this approach in outdoor environments. 
Similarly, \cite{yoon_adaptive_2024} present a method in which a robot’s previous experiences serve as inputs to a self-supervised learning system. This approach is particularly valuable in scenarios where no prior map is available, and traversability must be inferred from the robot’s past movements. Emphasizing the limitations of relying solely on immediate sensor input, \cite{cho_learning_2024} highlighted the importance of incorporating experiential data to enhance the accuracy of traversability estimation.

Despite significant advancements, the evaluation of traversability estimation methods is still an open issue, as there is no dedicated benchmark tailored for this task. Thus, many authors have generated their own labels in ad-hoc scenarios \cite{chavez-garcia_learning_2018}, \cite{seo_learning_2023}, which allow testing performance in real setups at the expense of reducing comparability and reproducibility. Others have opted to adapt already existing datasets containing semantic annotations of the scene \cite{cho_learning_2024}, \cite{yoon_adaptive_2024}. Benchmarks like Semantic-KITTI \cite{behley2019semantickitti} are typically used for structured environments. Traversability in the wild has been assessed instead through GOOSE \cite{mortimer_goose_2024} or RELLIS-3D \cite{jiang_rellis-3d_2021}. However, despite offering visual and semantic variety, they fall short in scenario diversity. More recently, NEGS-UGV \cite{sanchez_automatically_2022} presented a simulated dataset that provides a broader variety of terrain types and perfect 2D-3D semantic classes, making it a better fit for validating traversability estimation tasks.


\section{Method}

\label{sec:method}

The determination of traversable zones within the environment is performed by analyzing sensor data and odometry information collected during teleoperation. This process is formalized through the generation of feature vectors that describe the local properties of sub-gridmaps, as explained in this section.

Let \( G \in \mathbb{R}^{L \times N \times M} \) denote the multi-layer gridmap, where \( L \) is the number of layers, \( N \) and \( M \) are the grid dimensions, and each layer \( l \) encodes a specific property such as color (\( l_c \)) or elevation (\( l_e \)), so that \( G^{l}_{i,j} \) denotes the value of property \( l \) at cell \((i,j)\).

The information stored in each cell is dynamically calculated by associating the data captured by the LiDAR and camera sensors. Concretely, 3D points are projected onto the image space, and the RGB value of the corresponding pixel is obtained. Then, the spatial and color values are transformed to the reference frame of the gridmap to update both the \( l_c \) and \( l_e \) layers.

\begin{figure}[!ht]
  \centering
  \includegraphics[width=\linewidth]{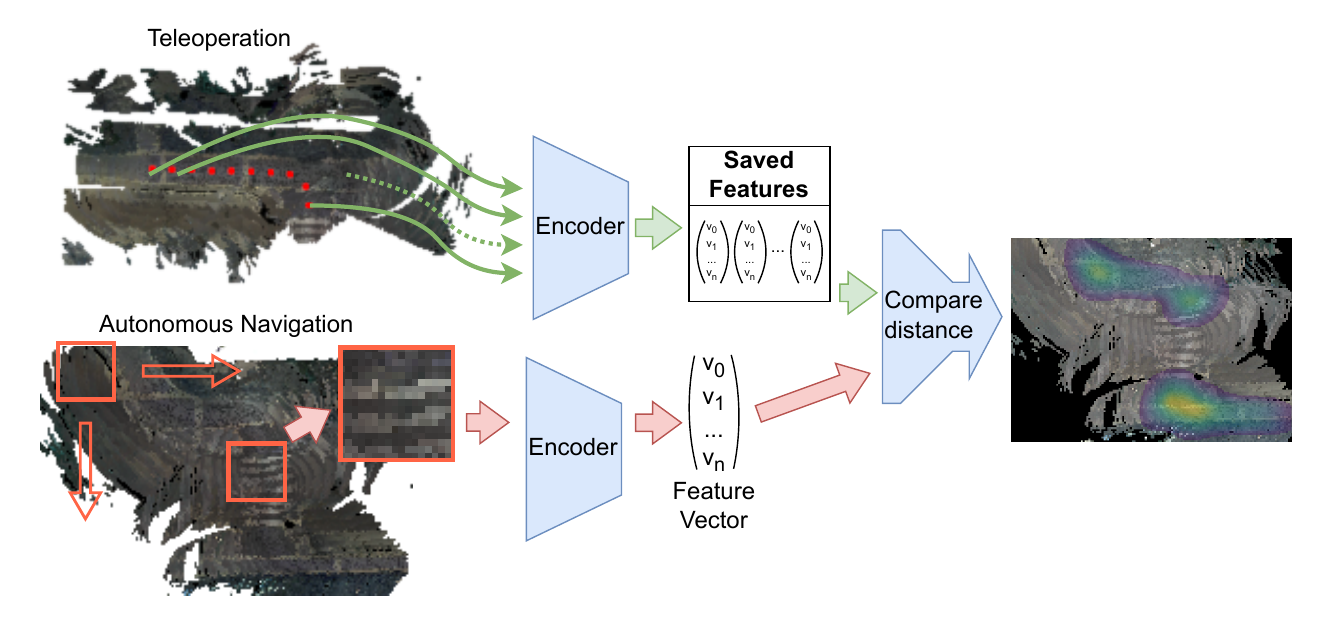}
  \caption{Method overview diagram}
  \label{fig:method_overview}
\end{figure}

Fig. \ref{fig:method_overview} illustrates a diagram that provides an overview of how our method functions, showing the process of feature vector extraction and comparison in the traversability assessment.

\subsection{Feature Vector Construction}
For each region of the gridmap that the robot has traversed, a sub-gridmap of size \( n \times n \) $G_t\subseteq G$ is extracted, and a feature vector $v_t = Features(G_t)$ is computed, which characterizes the properties of that specific region based on color and elevation of each cell. This feature vector is constructed using a variational autoencoder (VAE)\cite{kingma2013auto}.

The VAE-based approach allows for an abstract and compact representation of the region’s traversability characteristics, potentially improving generalization to unseen environments. In addition, VAE encoders favors the clustering of similar sub-gridmaps in nearby latent space locations.

Unlike other learning-based works in the literature 
\cite{frey_roadrunnerlearning_2024}, 
\cite{yoon_adaptive_2024}, 
\cite{zhang_learning_2024}, our proposed encoder does not require training with annotated data on the targeted navigation scenarios. Here, the VAE is trained on Describable Textures Dataset (DTD) \cite{cimpoi_describing_2014}, a dataset with multiple different textures and colors that enables learning rich feature vectors in a reduced latent space. 
VAE encoder follows a simple architecture of 3 convolutional layers that reduces the grid $16 \times 16 \times 4$ (3 RGB channels plus 1 height channel) to 32 features.

As the robot navigates through the environment, features vectors \( v_t \), are continuously extracted from each sub-grid and stored for future traversability assessments. However, to ensure scalability and efficiently manage the storage of feature vectors, we now employ the \textit{BIRCH} clustering algorithm \cite{zhang1996BIRCH}. \textit{BIRCH} incrementally clusters incoming vectors into compact subclusters, effectively maintaining representative summaries. If the new $v_t$ falls within an existing cluster radius (as defined by \textit{BIRCH}'s threshold parameter), it is considered redundant and not explicitly stored. Conversely, if $v_t$ is sufficiently distinct, \textit{BIRCH} creates or updates clusters accordingly, ensuring that only feature vectors representing significantly different regions of the environment are retained. This strategy significantly enhances scalability and storage efficiency while preserving critical information required for accurate traversability assessments.

\subsection{Traversability layer calculation}

The process of traversability evaluation is performed iteratively as shown in Algorithm \ref{alg:traversability} by sliding a window of size \( n \times n \) across the gridmap $G$. For each sub-gridmap $G_k$ within this window, a feature vector $v_t$ is extracted and compared to the \textit{BIRCH} model $\mathcal{B}$ sub-cluster centers to evaluate the traversability on that sub-gridmap. 

\begin{algorithm}[ht]
\caption{Traversability Gridmap Evaluation }
\label{alg:traversability}
\begin{algorithmic}[1]
\Require Gridmap $G$, \textit{BIRCH} model $\mathcal{B}$, window size $n$, threshold $\epsilon$
\State $G^{lT}_{i,j} \gets 0$ for all cells $(i,j)$ in $G$
\For{$c = 1$ to $\text{columns}(G) - n$}
    \For{$r = 1$ to $\text{rows}(G) - n$}
        \State $G_k \gets G_{c:c+n, r:r+n}$
        \State $v_t \gets Features(G_k)$

        \State $\text{cluster\_idx} \gets \mathcal{B}.\text{predict}(v_t)$
        \State $\text{cluster\_center} \gets \mathcal{B}.\text{subcluster\_centers}_{\text{cluster\_idx}}$
        \State $d \gets \|v_t - \text{cluster\_center}\|_2$
        \State $H \gets \text{get\_center\_weighted\_kernel}(\text{d})$
        \State $G^{lT}_{c:c+n, r:r+n} \gets \max\left(G^{lT}_{c:c+n, r:r+n}, H\right)$
    \EndFor
\EndFor
\State \Return{$G^{lT}$}
\end{algorithmic}
\end{algorithm}

The minimum euclidean distance $d$ between the calculated vector and the \textit{BIRCH} sub-cluster centers is used to assign a traversability score to the cells within the window, indicating how traversable the area is and allowing this value to be used as a cost in planning methods.

The window is then shifted across the gridmap, such that the process is repeated for multiple positions. Since the step of the sliding window is smaller than the window size, each cell in the gridmap may be evaluated multiple times. For each position, the computed traversability score is distributed across the window using a center-weighted traversability kernel, which assigns the highest value to the center cell and progressively lower values to surrounding cells as a function of their distance from the center. When a cell is evaluated by multiple windows, its final traversability score is set as the maximum between its current value and the new score computed for that window. This approach ensures that high traversability evidence from any window is preserved, providing a conservative yet informative estimate of the region’s traversability.


\section{Implementation}
\label{sec:implementation}

In this section, we present the implementation details and how we brought our approach into practice. To guarantee the reproducibility of our work, which is essential in any scientific work, we provide an open-source implementation that is fully reproducible using ROS 2 Jazzy. The project repository
\footnote{\href{https://github.com/IntelligentRoboticsLabs/global\_navigation}{https://github.com/IntelligentRoboticsLabs/global\_navigation}}
contains the necessary code and the rosbags used for the experiments, along with detailed instructions to facilitate replication.

We selected Gridmaps~\cite{fankhauser_universal_2016} to represent environmental knowledge and encode terrain traversability. This representation consists of multiple layers, allowing for the simultaneous depiction of elevation, occupancy, color, traversability, and other relevant data at each coordinate. The correspondence with real-world metric coordinates is determined by the resolution of each cell (typically 5-30 centimeters for most outdoor applications) and the offset of the grid's origin relative to the defined reference axis in the environment.


The choice of Gridmaps is also supported by the availability of an open-source implementation\footnote{\href{https://github.com/ANYbotics/grid\_map}{https://github.com/ANYbotics/grid\_map}}, which includes an API offering a variety of gridmap operations, such as extracting subgridmaps (patches). This implementation integrates well with ROS, making it highly adaptable for our application. The integration of this approach with Nav2, the most extended open-source navigation framework for ROS 2, is straightforward. 
Following \cite{rico_open_2024}, we utilized an Extended Map Server\footnote{\href{https://github.com/navigation-gridmap/extended\_map\_server}{https://github.com/.../extended\_map\_server}} capable of reading gridmaps and publishing them as an OccupancyGrid, which is the format Nav2 uses for maps. Our approach directly maps the terrain’s traversability onto the occupancy grid used by Nav2 for route planning.

\section{Experimental Validation}
\label{sec:exps}
To assess the performance of the proposed method, we have conducted a two-fold evaluation. First, we made use of the publicly available dataset NEGS-UGV \cite{sanchez_automatically_2022}, composed of a set of simulated sequences of a ground vehicle instrumented with camera and LiDAR sensors moving in different natural environments. This benchmark provides a perfect ground truth with pixel-wise semantic annotations and exact robot poses along the moving trajectories, making it ideal for evaluating traversability algorithms. 

It should be noted that formal evaluation is challenging since the ground truth in existing datasets is fixed (traversable/non-traversable), while our method dynamically evolves as the robot explores and learns new terrains. Consequently, standard metrics based on static annotations may not fully reflect the adaptive capabilities of our approach.

Additionally, the presented method was deployed into two different robotics platforms which were operated in real-world challenging scenarios. These scenes, precisely labeled for evaluation purposes, allow us to obtain quantitative results and analyze the ability of our work to adapt its output as the robot moves through new terrains.  


To perform direct comparisons against existing methods, we used a binary traversability classification by applying a threshold to pixel-wise predictions, categorizing them into traversable and non-traversable regions. For this comparative analysis, the $f_{0.5}$ score was employed as the evaluation metric. The $f_{0.5}$ score places greater emphasis on precision over recall, making it suitable for traversability and obstacle detection evaluations where precision is critical. This weighting is particularly important since false positives, or incorrectly classified traversable areas, are critically harmful. Navigating through areas falsely labeled as traversable can severely damage or immobilize the robot, thus precision becomes paramount in operational safety. The $f_{0.5}$ score is defined as:
\begin{equation}
f_{0.5} = (1 + 0.5^2) \cdot \frac{\text{Precision} \cdot \text{Recall}}{0.5^2 \cdot \text{Precision} + \text{Recall}}
\end{equation}

where Precision and Recall are computed by comparing each predicted image with the ground truth mask.

The ground truth for these evaluations is obtained from the dataset’s semantic point cloud. This cloud is processed by projecting the class labels of the points onto a grid map to generate the ground truth map. It is crucial to highlight that the proposed method is not a semantic segmentation algorithm. Its principal objective is not to detect predefined classes. Instead, the method focuses on identifying differences between regions, regardless of their semantic labels. Consequently, the regions identified as traversable or non-traversable by our method may encompass multiple semantic classes or sub-classes. 


Consequently, although precision-recall metrics offer insight into the method's capacity to differentiate between regions of varying traversability, direct comparison with results obtained from traditional semantic segmentation algorithms, which rely on a fixed number of classes, must be carefully interpreted. While the evaluation is performed against a segmented map with predefined classes, it is important to note that our method learns from a much broader set of classes. This allows it to be more critical when assessing whether an area is traversable or not, as will be demonstrated along this section.

To evaluate the effectiveness of our proposed approach, we conducted a comparative analysis against two state-of-the-art baselines, each representing a distinct and complementary methodology for traversability estimation:
\begin{itemize}
    \item \textbf{WVN} \cite{frey_fast_2023},
    a vision-based model that employs visual transformers on monocular RGB images. WVN is trained in a self-supervised manner using online feedback from proprioceptive sensors, allowing it to adapt to new environments during deployment. 
    \item \textbf{NAEX} \cite{Agishev2022}, 
    a LiDAR-based network designed for traversability segmentation. NAEX leverages the geometric properties of 3D point clouds to identify terrain characteristics, serving as a crucial component within a larger trajectory optimization pipeline. 
\end{itemize}
These two methods were selected not only for their strong performance but also for their contrasting modalities and strengths: WVN captures visual semantics through 2D imagery and online learning, while NAEX exploits 3D geometric features derived from LiDAR data. Together, they offer a comprehensive benchmark for assessing our approach's ability to generalize across different sensory modalities and terrain representations.





\subsection{NEGS-UGV Benchmark results}
The \textit{Natural Environment Gazebo Simulation of a Unmanned Ground Vehicle} (NEGS-UGV) \cite{sanchez_automatically_2022} dataset comprises a collection of four scenarios with uneven and diverse terrains such as grass, sandy shores, trails, or lakes. The variety of elements in the synthetic worlds also includes natural obstacles easily found while navigating in the wild, like trees, stones, or fallen trunks, as well as human-made objects that need to be avoided by a moving robot in operation (e.g. lamp posts).

\begin{figure}[ht]
     \includegraphics[width=\linewidth]{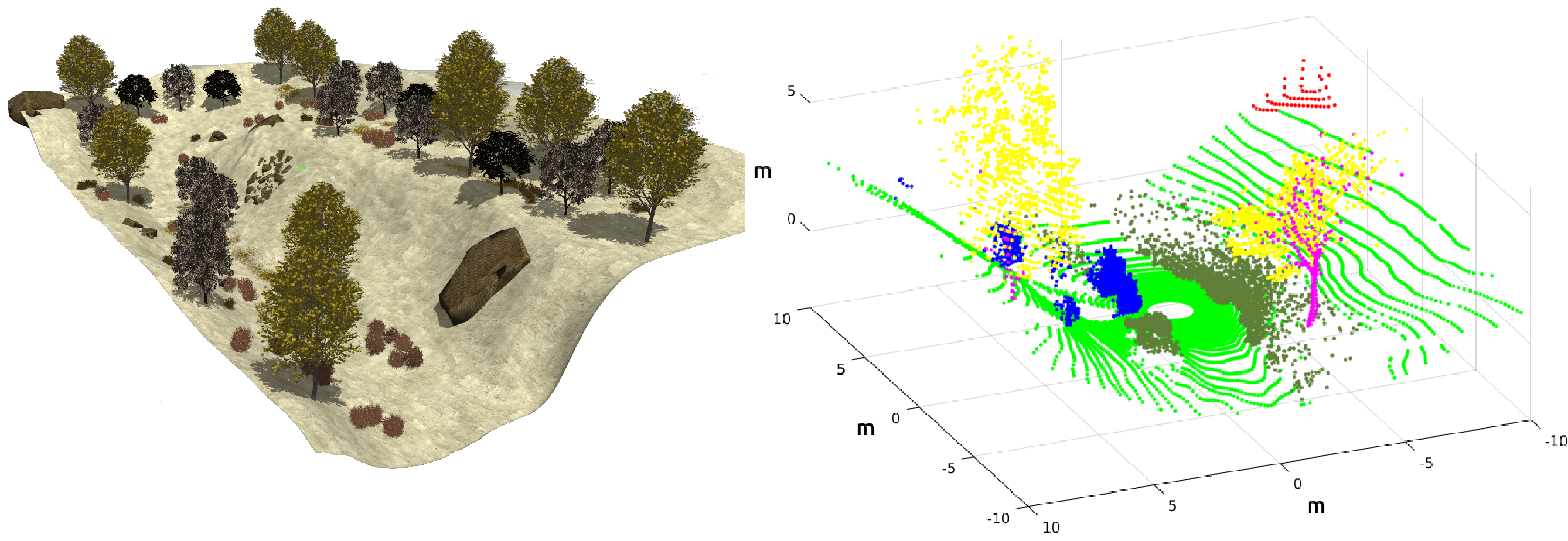}\label{negs_color}
    \caption{Sample frame from the NEGS-UGV dataset \cite{sanchez_automatically_2022} and its corresponding point semantic annotation.}
    \label{fig:negs-dataset}
\end{figure}

The benchmark was recorded using a model of a Husky robot equipped with a stereo camera, a LiDAR device, and a bundle of sensors dedicated to measuring the movement of the platform. As a simulated environment, exact robot poses and ground truth semantic labels are provided for both exteroceptive sensors. A sample frame from the dataset and its corresponding annotation is shown in Fig.~\ref{fig:negs-dataset}.


The ground truth assigns class-specific labels for every 3D point and every pixel in the image, discriminating between terrain kinds and vertical objects. 

This situation, convenient for validating traditional semantic segmentation tasks, differs from the traversability problem formulation studied in this work, which states that any terrain can be navigated by a robot as long as it has navigated through a similar area before.

Consequently, to evaluate the presented algorithm, the semantic classes need to be transformed into two categories: \textit{traversable} and \textit{non-traversable}. This classification is not trivial as, in our approach, the navigable area evolves according to the movement of the robot. 

To do so, it is key to take into consideration the path of the robot along the sequence, and adapt the traversable set of classes to include those surfaces the robot has walked through. Hence, it is appropriate to analyze the performance of the method over time.

To ensure a fair comparison with state-of-the-art methods that rely on a single sensor modality (either camera or LiDAR), traversability evaluation is performed in the map space. Semantic labels from all frames within the scene are aggregated based on the robot’s known positions. The 3D environment is then projected into a 2D horizontal map by quantizing the space into vertical pillars, with each cell representing the points contained within a corresponding pillar. Ground truth labels for this map are derived by categorizing each cell as either \textit{traversable} or \textit{non-traversable}. In cases where a cell includes points from both traversable and non-traversable categories (e.g., ground and a lamp post), the cell is conservatively labeled as \textit{non-traversable}.

\begin{figure*}[ht]
    \centering
\includegraphics[width=\linewidth]{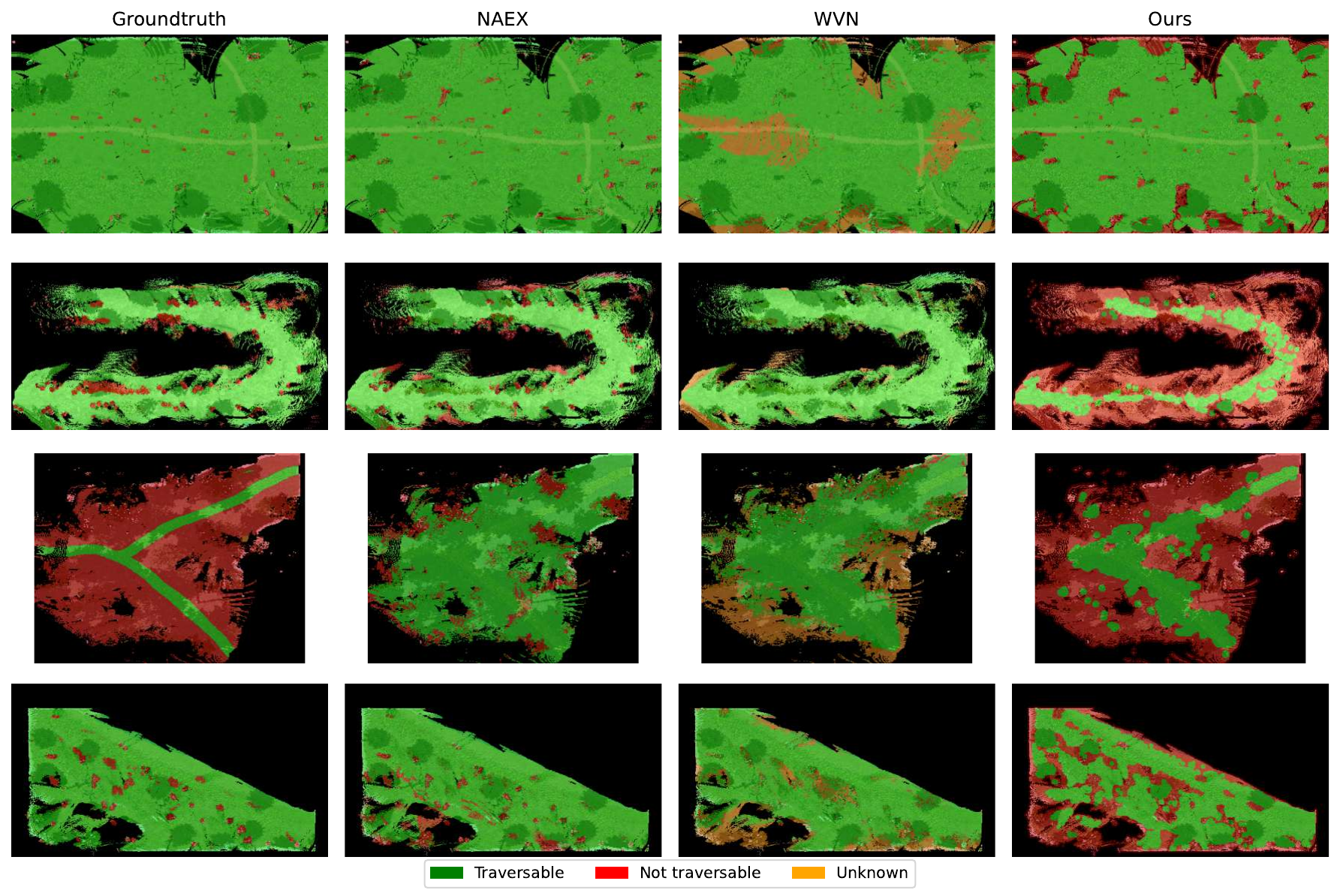}
    \caption{Estimated traversability map of the proposed method and state-of-the-art approaches in the literature in the four simulated scenarios of the NEGS-UGV dataset.}
    \label{fig:sota_negs_comparison}
\end{figure*}


To assess the impact of grid resolution on the performance of our method, we conducted an ablation study focused on the size of the subgridmaps used to extract features. The spatial resolution of the grid directly affects the area covered by each patch and thus the contextual information available for classification. Given that optimal feature extraction should reflect the robot’s physical footprint, we experimented with grid cell sizes of 0.05m, 0.1m, and 0.2m. For a $16 \times 16$ window size, these correspond to physical areas of approximately 0.8m, 1.6m, and 3.2m respectively, which are comparable to the size of the HUSKY robot employed in the NEGS-UGV dataset. As shown in Tab. \ref{tab:ablation_study}, the grid resolution of 0.1m yielded the best overall performance, striking a balance between contextual richness and spatial precision. Therefore, all subsequent experiments were conducted using a grid cell size of 0.1m.

\begin{table}[h]
\centering
\caption{Mean $f_{0.5}$ score across all scenarios for each tested grid cell size (ablation study).}
\label{tab:ablation_study}
\begin{tabular}{cc}
\toprule
Grid cell size (m) & Mean $f_{0.5}$ score \\
\midrule
0.05  & 0.74 \\
0.10  & 0.75 \\
0.20  & 0.67 \\
\bottomrule
\end{tabular}
\end{table}

In Fig. \ref{fig:sota_negs_comparison}, we present visual results of the generated maps for each scenario (Park, Hill, Forest, and Lake, arranged from top to bottom). Each row illustrates a distinct scenario, while each column corresponds, respectively, to the ground truth map, the comparative methods WVN and NAEX, and our proposed method. Additionally, Tab. \ref{tab:numeric_results} provides quantitative results in terms of the $f_{0.5}$ score for each experiment.

\begin{table}[h]
\centering
\caption{f0.5-score for each method in all four scenarios.}
\begin{threeparttable}
\begin{tabular}{lcccc}
\toprule
Sequence & NAEX & WVN & Ours & Ours\tnote{1} \\
\midrule
Park   & \textbf{0,999} & 0,998 &  0,97  & {} \\
Hill   & \textbf{0,957} & 0,941 &  0,2   & 0,757 \\
Forest & 0,178 & 0,271 & \textbf{0,475} &  {} \\
Lake   & \textbf{0,984} & 0,978 &  0,919 & {} \\
\bottomrule
\end{tabular}

\begin{tablenotes}
\item[1] Results obtained from a simulated trajectory designed to traverse sections of steep terrain.
\end{tablenotes}
\end{threeparttable}

\label{tab:numeric_results}
\end{table}



Analyzing the results, we observe that both NAEX and WVN perform better. This can be explained by the fact that almost the entire environment is traversable, which aligns to their optimistic behaviour observed in all the experiments. Even though, the difference compared to our method remains negligible, suggesting that, despite their better performance in these specific conditions, our approach is still equally effective.

Furthermore, in the Hill scenario, which, as depicted in Fig. \ref{fig:negs-dataset}, features steep slopes on both sides of the path, our method demonstrates more conservative criteria by only labeling the actual traversed path as navigable. This stricter classification occurs because the robot never navigates through highly inclined regions. Conversely, the ground truth map does not discriminate between these regions. Consequently, our metrics in this scenario initially appear substantially worse compared to those of competing methods.

To enable a more realistic evaluation and better demonstrate the capabilities of our approach, we simulated an alternative trajectory in the Hill scenario that explicitly includes navigation through the steep lateral areas. This additional exploration allows the robot to observe and learn that these regions are indeed navigable, leading to an improved traversability map. As shown in Table~\ref{tab:numeric_results}, the updated evaluation reflects a higher metric, indicating that a larger portion of the scenario is now correctly identified as traversable.

However, even with this enhanced trajectory, our method does not yet achieve the same metric values as those reported by other approaches. This discrepancy is primarily due to the complex and uneven terrain present in the Hill scenario. While our method can capture subtle variations in terrain geometry, the ground truth annotations lack the granularity to reflect this variability. Additionally, the simulated trajectory, though improved, still avoids highly sloped regions, limiting the method’s exposure to extreme terrain and, therefore, constraining overall performance.


The Forest scenario poses significant challenges due to its complex environment, characterized by flat terrain with subtle navigability distinctions heavily influenced by surfaces colour and trees shadows, complicating terrain interpretation. Nevertheless, our method demonstrates clear effectiveness by identifying and labeling the actual path as traversable. In contrast, WVN and NAEX, fail to extract these crucial path characteristics and incorrectly classify the majority of the environment as navigable. This clear difference emphasizes that our method excels in more complex settings, where minimal differences in geometry and texture are key to distinguish traversable and non-traversable areas.

\subsection{Real-world scenarios results}

In this section, we present the results from experiments conducted using two real robotic platforms, shown in Fig. \ref{fig:robots}:
\begin{itemize}
    \item \textbf{Robotnik RB-Summit}: Powered by both an Intel NUC and a Jetson Xavier AGX, each running Ubuntu 18. This robot mounts a 16-plane 3D LiDAR and a Zed2 camera.
    \item \textbf{Unitree Go2}: Equipped with a Jetson Orin NX running Ubuntu 20, an ultra-wide 4D LiDAR, a 32-plane 3D LiDAR, and an Intel RealSense 435i camera.
\end{itemize}

The proposed method was validated in two distinct scenarios where the robot was teleoperated. The environments tested were a grassy path and a sidewalk with a crosswalk. 


\begin{figure}[htb]
  \centering
  \includegraphics[width=\linewidth]{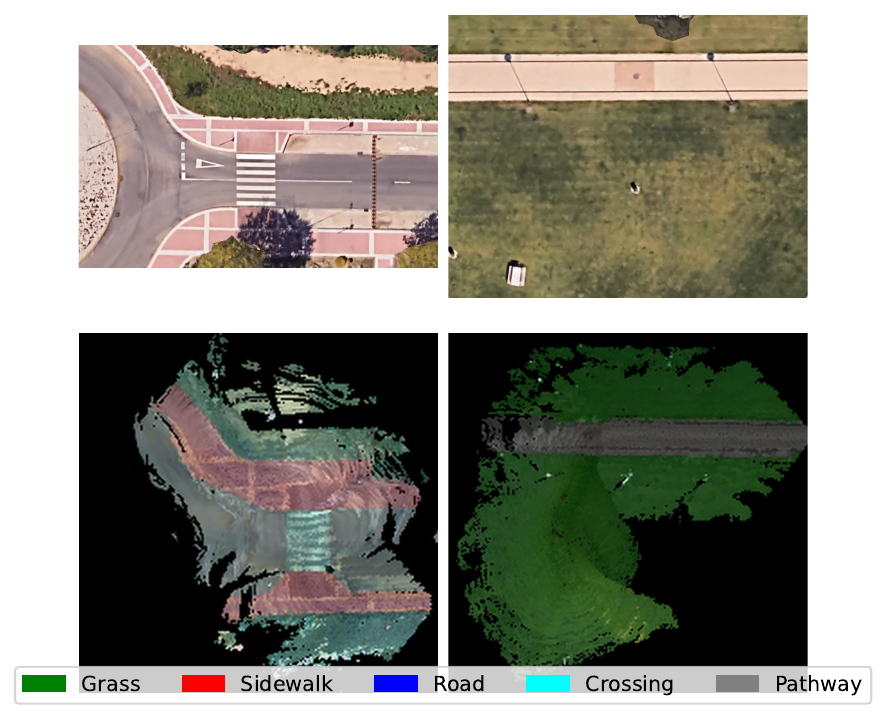}
  \caption{Top: satellite views of the two experimental environments. Bottom: corresponding manual semantic annotations of those same areas.}
  \label{fig:gt_scenarios}
\end{figure}

To ensure a wide range of terrain representation, the trajectories were designed to include specific terrain types: 

The first scenario assesses the traversability in environments comprising paved pedestrian pathways and grassy areas (Fig. \ref{fig:gt_scenarios} right). Certain robots, such as the Go2, can navigate through terrains challenging for wheeled vehicles, like grass. Initially, the robot only navigates through the paved pathway, but eventually, the robot leaves the pathway and starts to navigate on the grass. In this experiment, according to the definition of traversability adopted in this work, only the pathway should be considered traversable until the robot crosses the grass, at which point the grass should also be considered traversable.

The second scenario involves a sidewalk and crosswalk setting (Fig. \ref{fig:gt_scenarios} left). Initially, the robot navigates exclusively on the sidewalk. The expected behavior of the robot at this point is to mark as navigable, not only the sidewalk on which it has navigated, but also the one on the other side. After that, the robot crosses the crosswalk, whereupon it should mark as navigable also the crosswalk, but not the road.

Table \ref{tab:exp_results} presents the results for the two real-world scenarios (S1 and S2) at different time points. These time instants correspond to stages in the robot's exploration where its understanding of traversability has changed, resulting in different terrain classes being recognized as navigable.

\begin{table}[ht]
\centering
\caption{F0.5-scores in real-world scenarios (S1, S2) at two instants, reflecting evolving traversable areas. "Classes" list the terrain types navigated by the robot at each instant.}
\label{tab:f1_scores}
\begin{tabular}{llccc}
\toprule
Scenario      & Classes              & NAEX & WVN & Ours \\
\midrule
S1            & Pathway                 & 0.20    & 0.37    & \textbf{0.72} \\
S1            & Pathway, Grass          & \textbf{0.99} & 0.95    & 0.92    \\
S2            & Sidewalk             & 0.37    & 0.50    & \textbf{0.63} \\
S2            & Sidewalk, Crossing   & 0.43    & 0.58 & \textbf{0.63}    \\
\bottomrule
\label{tab:exp_results}
\end{tabular}
\end{table}

As can be observed, the proposed method demonstrates a superior ability to distinguish among different types of terrain, resulting in more precise decisions regarding traversability. 


For example, the GO2 quadruple might successfully traverse grass-covered areas, whereas the Summit wheeled robot could find the same terrain impassable. As reflected in our results, our method achieves notably higher metrics when the robot has only previously encountered a single class of terrain (e.g., path or sidewalk) and has no prior exposure to other classes. However, as the robot's capability to navigate multiple types of terrain increases, the performance metrics of our method converge to those of the baseline methods, which inherently exhibit limited differentiation and generally label broad areas as traversable by default.


\begin{figure}[htb]
  \centering
  \includegraphics[width=0.8\linewidth]{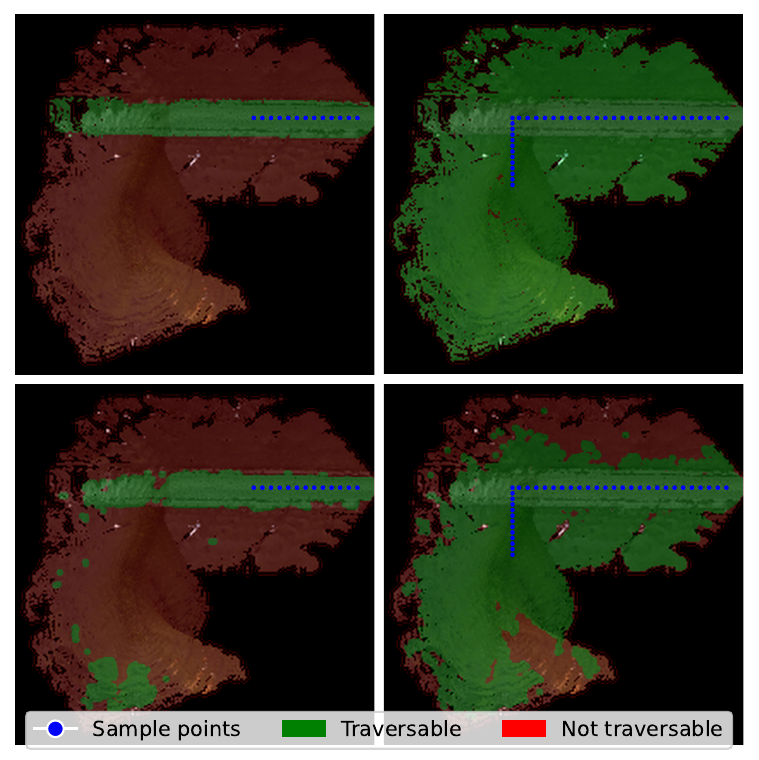}
  \caption{Comparison of annotated versus predicted traversability maps. Top row: ground truth from robot trajectories, where left only considers pathway and right includes also grass. Bottom row: proposed method outputs. Sample points are overlaid.}
  \label{fig:pisandocesped}
\end{figure}

Our method's adaptive learning capabilities are further illustrated in Fig. \ref{fig:pisandocesped}, where it can be observed how, after traversing only a limited segment of the path, the algorithm accurately generalizes, identifying the entire pathway as navigable (left). Furthermore, once the robot encounters the grass area, the method promptly learns to recognize the grass surface as traversable, thus dynamically expanding the region classified as navigable (right). This shows the effectiveness of our approach in incrementally updating terrain knowledge based on direct interaction.


\section{conclusions}
\label{sec:Conclusions}

In this work, we proposed a novel method for traversability estimation in field robotics that enables a robot to identify navigable terrain based solely on its prior experiences, without requiring large-scale semantic annotations or predefined terrain definitions. This learning-by-demonstration paradigm simplifies the deployment of autonomous robots in unfamiliar environments.

Our approach fuses LiDAR and RGB data to build a multi-layer grid representation of the environment. This representation is processed by a lightweight VAE to generate a compact and informative encoding for each map region. These feature vectors are then compared against previously learned traversable regions to assign a traversability score to each cell.

We validated the method in both synthetic and real-world scenarios. Experiments using the NEGS-UGV benchmark demonstrated that our method achieves comparable or superior performance to state-of-the-art approaches, particularly in diverse and restrictive terrains. Furthermore, tests with multiple robot platforms confirmed the robustness and platform-agnostic nature of the proposed methodology, highlighting its capacity to generalize across terrains and robot morphologies.

Future work will explore enriching the input representation with features such as surface reflectivity, estimated friction, and stiffness derived from both exteroceptive and proprioceptive data to allow the VAE to learn more discriminative features and improve the accuracy and generalization of traversability estimation.

\bibliographystyle{IEEEtranS}
\bibliography{IEEEabrv, references_v3}
 
\end{document}